\documentclass[
]{ceurart}

\usepackage{cleveref}

\begin{document}

\copyrightyear{2022}
\copyrightclause{Copyright for this paper by its authors.
  Use permitted under Creative Commons License Attribution 4.0
  International (CC BY 4.0).}

\conference{ITADATA 2022: The $1^{st}$ Italian Conference on Big Data and Data Science,
  September 20--21, 2022, Milan, Italy}

\title{Benchmarking FedAvg and FedCurv for Image Classification Tasks}

\author[1]{Bruno Casella}[%
orcid=0000-0002-9513-6087,
email=bruno.casella@unito.it,
url=https://alpha.di.unito.it/bruno-casella/,
]

\author[1]{Roberto Esposito}[%
orcid=0000-0001-5366-292X,
email=roberto.esposito@unito.it,
url=https://www.unito.it/persone/esposito,
]

\author[2]{Carlo Cavazzoni}[%
orcid=0000-0002-9589-4785,
email=carlo.cavazzoni@leonardo.com,
]

\author[1]{Marco Aldinucci}[%
orcid=0000-0001-8788-0829,
email=marco.aldinucci@unito.it,
url=https://alpha.di.unito.it/marco-aldinucci/,
]

\address[1]{Department of Computer Science, University of Torino, Torino, Italy}
\address[2]{Leonardo S.p.A., Italy}
  
\begin{abstract}
Classic Machine Learning (ML) techniques require training on data available in a single data lake (either centralized or distributed). However, aggregating data from different owners is not always convenient for different reasons, including security, privacy and secrecy.  
Data carry a value that might vanish when shared with others; the ability to avoid sharing the data enables industrial applications where security and privacy are of paramount importance, making it possible to train global models by implementing only local policies which can be run independently and even on air-gapped data centres. 
Federated Learning (FL) is a distributed machine learning approach which has emerged as an effective way to address privacy concerns by only sharing local AI models while keeping the data decentralized. 
Two critical challenges of Federated Learning are managing the heterogeneous systems in the same federated network and dealing with real data, which are often not independently and identically distributed (non-IID) among the clients. In this paper, we focus on the second problem, i.e., the problem of statistical heterogeneity of the data in the same federated network. In this setting, local models might be strayed far from the local optimum of the complete dataset, thus possibly hindering the convergence of the federated model. Several Federated Learning algorithms, such as FedAvg, FedProx and Federated Curvature (FedCurv), aiming at tackling the non-IID setting, have already been proposed. This work provides an empirical assessment of the behaviour of FedAvg and FedCurv in common non-IID scenarios. Results show that the number of epochs per round is an important hyper-parameter that, when tuned appropriately, can lead to significant performance gains while reducing the communication cost. As a side product of this work, we release the non-IID version of the datasets we used so to facilitate further comparisons from the FL community. 
\end{abstract}


\maketitle

\section{Introduction}
\label{sec:introduction}
The increasing availability of big data and computational resources supports the growth of accurate and reliable Machine Learning (ML) and Deep Learning (DL) models. More and more sectors, from the public to companies, follow data-driven approaches in their everyday decisions. One of the difficulties in building systems that support these decisions is the necessity to obtain large datasets for training the models. 
In fact, in many situations, data is scattered between many parties, and it would be beneficial to merge it into a single place in order to gather the information needed to build high-quality models. 
However, due to increasing privacy as well as security concerns merging the data might not be a feasible approach. For instance, the European GDPR ~\cite{voigt2017eu} regulation adds strong constraints on the possibility of sharing data between parties when sensible pieces of information are at stake. 
Moreover, industrial data are often not shared even when allowed by the law because they represent an essential advantage over competitors or because of the security concerns deriving from the need to expose part of the data to other parties. 
Consequently, many players resort to train their models on local datasets, and the resulting AI models often suffer from low reliability, leading to generalizability and overfitting issues ~\cite{zech2018variable}. 

Federated Learning (FL) ~\cite{mcmahan2017communication} is a machine learning setting that emerged as an effective way to train on distributed datasets without requiring any exchange of data-related information between the parties. 
This not only solves the privacy concerns, but also allows the parties to implement local policies that can be deployed more securely even in extreme settings such those required by air gapped systems.

In a typical FL scenario, at each round, multiple clients take one or more steps of gradient descent on a shared model using their local data, and then a central node acts as an aggregator performing a weighted average (Federated Averaging ~\cite{mcmahan2017communication}) of the resulting models. This approach generally performs well when the training data distributions are independent and identically distributed (IID). However, in a real-world scenario, due to the natural differences in the collection of data, FL faces difficulties due to data being non-independent and identically distributed (non-IID) among the involved parties. In these cases, the data of a single client does not represent the overall distribution of data among the federation, and this poses a key challenge for FL ~\cite{kairouz2021advances}. 
For example, in image recognition, models trained on sunny days may not be accurate on cloudy days due to an unrecognizable feature distribution.

The first and most used FL algorithm is FedAvg, proposed by Google in 2017, which can actually work on non-IID data when strong assumptions hold on to the loss function. Specifically, a recent work \cite{li2019convergence} shows that if the optimization problem is strongly convex and smooth, then FedAvg would converge even non-IID data. FedAvg averages local gradients of the nodes of the federation. Recent years have seen the deployment of several FL algorithms for addressing the non-IID setting, like FedProx ~\cite{li2020federated}, FedNova ~\cite{wang2020tackling} and SCAFFOLD ~\cite{karimireddy2020scaffold}. All these algorithms have already been tested in \cite{li2022federated}, which provides valuable information about their behaviour in non-IID settings. In this paper, we compare the performance of FedAvg with FedCurv ~\cite{shoham2019overcoming} on several common non-IID settings. FedCurv is an algorithm that addresses the problem of catastrophic forgetting ~\cite{goodfellow2014empirical} in FL by adding a penalty term to the loss function in order to compel the local models to a shared optimum. We tested both algorithms on three public image datasets, i.e. MNIST ~\cite{lecun1998gradient}, CIFAR10 ~\cite{krizhevsky2009learning}, and MedMNIST ~\cite{medmnistv2}. We manipulate the dataset to simulate common non-IID settings and show that performance is often related to the number of epochs performed locally in each round of the FL algorithm. 

The main contributions of this work are:
\begin{itemize}
    \item we provide a benchmark datasets for five different non-IID settings (see Section~\ref{sec:non-iid-data}): quantity skew, three versions of prior shift and covariate shift;
    \item we provide results of extensive experiments on FedAvg and FedCurv on the considered non-IID settings as well as on the IID case. To the best of our knowledge, this is the first work providing empirical evidence on the behaviour of FedCurv in common non-IID cases;
    \item we discuss the results under several points of view, thus providing a faceted understanding of the behaviour of the algorithms;
    \item as a result of the experimentation, we show that the number of epochs per round can be pivotal in obtaining good performances while limiting the communication costs.

\end{itemize}


The rest of the paper is organized as follows. In Section~\ref{sec:related-work}, we present the recent related works. Section~\ref{sec:fl-algorithms-on-non-iid-data} reviews the FL algorithms tackling non-iidness. In Section~\ref{sec:experiments} are given some of the most typical non-IID partition strategies, and experimental results are shown and discussed. Finally, in Section~\ref{sec:conclusions}, conclusions are drawn.

\section{Related Work}
\label{sec:related-work}
Dealing with non-IID data represents one of the fundamental challenges in FL. \cite{kairouz2021advances} surveys this issue providing several non-IID data cases.

To the best of our knowledge, there are only a few benchmarks for FL in the non-IID scenario. One work ~\cite{liu2020evaluation} tests only FedAvg on MNIST and CIFAR10 in the quantity skew and labels quantity skew settings. Another work \cite{li2022federated} considers quantity skew, labels quantity skew and three types of feature distribution skew, i.e. noise-based, synthetic and real-world feature imbalance. It tests FedAvg, FedProx ~\cite{li2020federated}, FedNova ~\cite{wang2020tackling} and SCAFFOLD ~\cite{karimireddy2020scaffold} on 9 public image datasets, including MNIST and CIFAR10. This work points out how non-iidness represents a challenge for FL algorithms in terms of accuracy, showing how none of those algorithms outperforms others in all the tested cases. 

A more recent work ~\cite{polato2022boosting} proposes three FL algorithms based on distributed boosting strategies rather than gradient-descent based methods: \textit{DistBoost.F} and \textit{PreWeak.F}, that are the adaptation to the FL setting of the distributed boosting algorithms DistBoost \cite{lazarevic2002boosting} and PreWeak \cite{cooper2017improved}, and \textit{AdaBoost.F}, that is a novel algorithm developed for FL. The authors performed experiments on ten tabular UCI datasets, comparing their proposed algorithms in the same non-iid settings of this paper. This work shows that in most cases, the aggregated model outperforms the models that could have been learned on local data, but that non-iidness can hurt the efficiency of the federation.

Apart from these works, there are some benchmarks for FL that do not focus on non-IID data issues. LEAF \cite{caldas2018leaf} is a modular benchmarking framework for FL, which includes a set of both image and tabular datasets, an evaluation framework and a set of reference implementations. Some open-source datasets they provided are \textit{FEMNIST} (Federated Extended MNIST), which is built by partitioning the data in Extended MNIST \cite{cohen2017emnist} based on the writer of the digit/character, and \textit{Shakespeare}, a dataset built from \textit{The Complete Works of William Shakespeare} \cite{shakespeare2014complete}, where each speaking role in each play is considered a different device. The Open Application Repository for Federated Learning (OARF) \cite{hu2020oarf} is a benchmark for FL mimicking realistic scenarios with publicly available datasets as different data silos in image, text and structured data. FedML \cite{he2020fedml} is an open research library and benchmark developed for fair algorithm performance comparison. It provides algorithm development of FedAvg, FedNova and FedOpt \cite{reddi2021adaptive}.

\section{FL Algorithms on non-IID data}
\label{sec:fl-algorithms-on-non-iid-data}
In a non-IID setting, the local data of a single client can not represent the overall distribution of the federation. In such a situation, the local models drift apart because local optima may be far from the global optima. This results in a reduction of accuracy of both local models, which are driven towards the local optima, and of the aggregated model because local updates may be large, in particular when each round has a lot of local epochs. In such conditions, the aggregated model can have a worse accuracy than models learnt using only local data.

Several approaches have already been proposed to tackle the problem of non-iidness. The most popular algorithm for FL in the non-IID settings are FedAvg ~\cite{mcmahan2017communication}, FedCurv ~\cite{shoham2019overcoming}, FedProx ~\cite{li2020federated}, FedNova ~\cite{wang2020tackling} and SCAFFOLD ~\cite{karimireddy2020scaffold}. The main features of these algorithms are described below.

\noindent\textbf{FedAvg:}
Federated Averaging \cite{mcmahan2017communication} has been the first algorithm to be proposed for the FL setting. Basically, the shared model parameters are initialized by the aggregator at the beginning of the training. Afterwards, each client trains a local copy of the model on its local data and sends the result to the server, which sets the shared model to be the (weighted) average of the received local models. Then the aggregator sends the shared model again to each client, and the process repeats. Typically in cross-device settings, the server sends the global model to a random subset of clients in order to cope with the high number of parties in the federation, which can be in the order of thousands or even more. Two parameters that can be controlled and might have a large impact on the results are the number of local training epochs per round (E) and the local batch size (B). If $B=\infty$ and $E=1$ the algorithm is called FedSGD \cite{mcmahan2017communication}. With $E>1$, the number of communication rounds can decrease; however, local models can be driven towards the local optima, leading to bad accuracy. In \cite{li2019convergence} it is shown that if the loss is strongly convex and smooth, the rate of convergence of FedAvg is $O(1/T)$ (where $T$ is the number of rounds) and that weight decay is a necessary condition for optimal convergence. 

\noindent \textbf{FedCurv:}
Federated Curvature (FedCurv) \cite{shoham2019overcoming} builds on the idea of Lifelong Learning \cite{parisi2019continual} to prevent catastrophic forgetting \cite{goodfellow2014empirical} in FL. It is an adaptation for FL of  Elastic Weight Consolidation (EWC) \cite{kirkpatrick2016overcoming}, an algorithm for sequentially training a model on new tasks without forgetting old ones. The basic assumption of EWC is that neural networks are sufficiently over-parameterized that there are good chances of finding an optimal solution $B^*$ to task B in the neighbourhood of a solution $A^*$ learned on a previous task $A$. EWC uses the diagonal of the Fisher information matrix to choose the most important weights of the previous task. EWC adds a penalty term to the optimization function in order to force the model parameters with the higher Fisher information for task A to maintain their actual value while learning task B.
FedCurv adds this EWC penalty term for minimizing the model disparity across the clients of a federation. During each round, FedCurv works just like FedAvg, but each client sends its local model together with the diagonal of Fisher's information matrix. Since this method allows for less frequent communication 
fewer steps are required in order to reach the desired accuracy. However, in each round, the number of parameters to be transmitted is about three times that in FedAvg.

\noindent \textbf{FedProx:}
FedProx \cite{li2020federated} is a re-parametrization of FedAvg aiming to tackle two key challenges in FL: systems heterogeneity (variability of the devices) and statistical heterogeneity (non-iidness). 
FedProx alleviates systems heterogeneity starting from the idea of allowing a variable amount of work across devices based on their resource constraints. In the case of statistical heterogeneity, local models may drift apart. FedProx restricts the size of local updates by adding an L2 regularization term in the cost function compelling the local model to stay close to the global model. L2 regularization is controlled by an additional term $\mu$ that needs to be tuned carefully.
FedProx does not introduce communication overhead, but it increases the computation overhead of the devices. 

\noindent \textbf{FedNova:}
\label{ssec:fednova}
Federated Normalized Averaging (FedNova) \cite{wang2020tackling} proposes a slightly modified version of FedAvg to overcome the problem of objective inconsistency while preserving fast error convergence. The paradigm is the same as FedAvg, but FedNova normalizes and scales local updates of the parties based on their number of local steps (mini-batches of local training). 

\noindent \textbf{SCAFFOLD:}
Stochastic Controlled Averaging for federated learning (SCAFFOLD) \cite{karimireddy2020scaffold} proposes a solution for client-drift. 
SCAFFOLD requires significantly fewer communication rounds, however, it doubles the communication size per round due to the additional control variates.

\section{Non-IID data}
\label{sec:non-iid-data}
An existing study \cite{kairouz2021advances} identifies five different non-IID scenarios: 1) prior shift (label distribution skew), 2) covariate shift (feature distribution skew), 3) same label but different features, 4) same features but different labels, and 5) quantity skew. Many of these cases (1, 2 and 5) have already been tested in \cite{li2020federated}. In our work, we focus on the iid case (uniform data distribution) and on five types of non-iidness: quantity skew, three versions of prior shift and covariate shift (feature distribution skew). These scenarios have already been tested in \cite{polato2022boosting} with tabular datasets and non-gradient-descent methods. 
We briefly describe the adopted settings.

\noindent \textbf{Quantity Skew:} a collection of datasets exhibits a quantity skew if different clients (datasets) can hold vastly different amounts of data. In this case, the sampling distribution may still be consistent among the parties; it is, however, interesting to see the effect of the quantity imbalance on the convergence of the FL algorithm. 
In this case the proportion of example to be assigned to client $x \in \{0\dots N-1\}$ is determined by the Power distribution $P(x; \alpha) = \alpha x^{\alpha-1}$. Here $\alpha > 0$ is a parameter affecting the distribution as it follows: 
if $\alpha = 1$ then examples are uniformly  distributed across clients; if $\alpha = 2$ examples are "linearly" distributed across users (our case);
$if \alpha \geq 3$ the examples are power law distributed;
in general, as $\alpha \rightarrow \infty$, it is increasingly true that a single user obtains most of the examples, and the remaining users only get very few examples.

\noindent \textbf{Prior shift:} let us consider the conjunct distribution of data and labels $P(x_i, y_i) \ = P(x_i|y_i)P(y_i)$, a collection of datasets exhibit a prior shift if the labels' prior $P(y_i)$ varies across clients. 
This is a common scenario in many real-world FL applications. For instance, labels' prior can vary when devices are spread across different world regions, and certain elements are present only in some countries. The types of prior shift implemented are the following:
\begin{description}
    \item \textbf{Labels quantity skew:} this is the simplest version of prior shift. Labels are partitioned between clients and each client receives samples that belongs to only a fixed number of classes. In our experiments, we fixed the number of classes per client to 2.
    \item \textbf{Dirichlet labels skew:} in this case the number of examples of a given class that each client receives is distributed according to samples extracted from the Dirichlet distribution. More specifically, for each class $k$, we extract an $N$ dimensional vector $p_k$ from $\text{Dir}_N([\beta, \dots, \beta])$ and assign a proportion of $p_{k,j}$ samples of class $k$ to client $j$. The N-dimensional vector of $\beta$s is the \emph{concentration parameter}. The lower $\beta$, the greater the imbalance. In our experiments we fixed $\beta = 0.5$ as in \cite{polato2022boosting,li2020federated}.
    \item \textbf{Pathological labels skew:} this skewness was designed in \cite{mcmahan2017communication}. First of all, data are sorted by label and divided into shards of size $N \cdot shards\_per\_client$. Each client owns shards\_per\_client shards. We fixed $shards\_per\_client = 2$. Basically, in this case, most parties will only have a limited number of labels.
\end{description}

\noindent \textbf{Covariate shift: } in covariate shift, also known as distribution skew, the marginal distributions $P(x_i)$ may vary across clients, even if $P(y|x)$ is shared. It is common to encounter covariate shift in several fields of machine learning. For example, in speech recognition, a model trained in a specific language can have trouble when it encounters particular accents or dialects of that language; in image recognition, a model trained on sunny days may not be accurate on cloudy days due to an unrecognizable feature distribution. To simulate covariate shift on our datasets, we followed the procedure outlined in \cite{polato2022boosting}, where examples are assigned to clients according to a distribution based on the results of a Principal Component Analysis (PCA) performed on the data.

\section{Experiments}
\label{sec:experiments}
To conduct our experiments, we adopted Open Federated Learning (OpenFL) \cite{reina2021openfl}, the new framework for FL developed by the Intel Internet of Things Group (IOTG) and Intel Labs. OpenFL is a Python 3 library for FL that is Deep Learning framework-agnostic. Training of statistical models may be done with any deep learning framework, such as TensorFlow or PyTorch, via a plugin mechanism. OpenFL employs an Aggregator-Collaborator workflow: a Collaborator is a client in the federation that trains a global model on a local dataset, while the Aggregator aggregates the model updates received from Collaborators. All the experiments were computed in a distributed environment encompassing $N=10$ collaborators. Each collaborator is run on an Intel Xeon CPU (8 cores per CPU).

\noindent\textbf{Dataset:} We compared FedAvg and FedCurv on MNIST \cite{lecun1998gradient}, CIFAR10 \cite{krizhevsky2009learning} and MedMNIST \cite{medmnistv2}. CIFAR10 and MNIST are default benchmarks in NN literature; MedMNIST has been selected due to the increasing interest of DL and FL in medical domains. In particular, we used OrganAMNIST, a 2D dataset on Abdominal CT contained in MedMNIST. The details of the datasets are summarized in Table I.

\begin{table}
\caption{\label{tab:dataset-stats}Statistics of the datasets.\\}
\label{tab:freq}
\begin{tabular}{llll}
\toprule
\textbf{Dataset}  & \textbf{Train samples} & \textbf{Test samples} &  \textbf{\# labels} \\
\midrule
MNIST    & 60.000        & 10.000       & 10        \\
\midrule
CIFAR10  & 50.000        & 10.000       & 10        \\
\midrule
MedMNIST & 34.581        & 17.778       & 11        \\ 
\bottomrule
\end{tabular}
\end{table}

\noindent\textbf{Preprocessing:} 2D datasets MNIST and MedMNIST were kept at 28x28, while CIFAR10 was rescaled to 64x64. As for data augmentation, we performed random horizontal flips and angle rotation of 10° with a probability of 80\%. All the datasets were normalized according to their mean and standard deviation. 

\noindent\textbf{Model:} We employed ResNet-18~\cite{he2016deep} as classification model, trained by minimizing the cross-entropy loss with mini-batch gradient descent using the Adam optimizer with learning rate $10^{-4}$. The local batch size was 64.

The algorithms have been compared using the standard classification metric of top-1 accuracy.
Results are reported in  \cref{tab:non-fed-setting,tab:uniform-setting,tab:quantity-skew,tab:prior-shift,tab:covariate-shift}, with \cref{tab:non-fed-setting} reporting about a non-federated baseline, i.e., the typical AI scenario where the data are aggregated in a single device. The remaining tables reports the results about experiments in different non-iid settings of the two tested federated algorithms.


\subsection{Discussion}
\label{ssec:discussion}

\begin{table}
\caption{\label{tab:non-fed-setting}Classification accuracy in the non-federated setting.\\}
\label{fig:non-federated}
\begin{tabular}{lrc} 
\toprule
\textbf{Datasets}           & \textbf{10 epochs}   & \textbf{100 epochs}   \\ 
\midrule
MNIST       & 98.67$\%$               & 99.12$\%$         \\
\midrule
CIFAR10       & 64.03$\%$               & 71.25$\%$         \\
\midrule
MedMNIST       & 84.85$\%$               & 89.13$\%$         \\
\bottomrule
\end{tabular}
\end{table}

\begin{table}
\caption{\label{tab:uniform-setting}Comparison between FedAvg and FedCurv in the uniform setting.\\}
\label{fig:uniform}
\begin{tabular}{lrcccc} 
\toprule
\multicolumn{1}{l}{}   &                      & \multicolumn{2}{c}{\textbf{10 rounds}}                                                               & \multicolumn{2}{c}{\textbf{100 rounds}}                                                            \\ 
\cmidrule(lr){3-4} \cmidrule(lr){5-6}
\textbf{Datasets}      &  \textbf{Epochs}                     & \textbf{FedAvg} & \textbf{FedCurv}  & \textbf{FedAvg} & \textbf{FedCurv}  \\ 
\midrule
\multirow{3}{*}{MNIST}        & 1                    & 97.16$\%$               & \textbf{97.17$\%$}     & \textbf{97.66$\%$}      & 97.53$\%$    \\
                           & 10                   & 99.05$\%$               & \textbf{99.07$\%$}     & \textbf{99.35$\%$}    &   99.33$\%$     \\
                           & 30                  & 99.28$\%$              & \textbf{99.33$\%$}     & 99.47$\%$      & \textbf{99.55$\%$}             \\ 
\midrule
\multirow{3}{*}{CIFAR10}        & 1                    & \textbf{56.34$\%$}               & 56.13$\%$     & 5\textbf{4.89$\%$}      & 54.78$\%$    \\
                           & 10                   & \textbf{70.95$\%$}               & 70.40$\%$     & 74.03$\%$      & \textbf{75.57$\%$}     \\
                           & 30                  & 74.05$\%$              & \textbf{74.07$\%$}     & \textbf{78.89$\%$}      & 78.57$\%$             \\ 
\midrule
\multirow{3}{*}{MedMNIST}        & 1                    & 68.98$\%$               & \textbf{79.70$\%$}     & 72.42$\%$      & \textbf{83.50$\%$}    \\
                           & 10                   & 68.19$\%$               & \textbf{86.90$\%$}     & 72.96$\%$      & \textbf{89.49$\%$}     \\
                           & 30                  & 46.00$\%$              & \textbf{88.77$\%$}     & 71.16$\%$      & \textbf{90.23$\%$}             \\ 
\midrule
\# best performance  & & 2 & 7 & 4 & 5 \\
\bottomrule
\end{tabular}
\end{table}

\begin{table}
\caption{\label{tab:quantity-skew}Comparison between FedAvg and FedCurv in the quantity skew setting.\\}
\label{fig:quantityskew}
\begin{tabular}{crcccc} 
\toprule
\multicolumn{1}{l}{}       &                      & \multicolumn{2}{c}{\textbf{10 rounds}}                                                               & \multicolumn{2}{c}{\textbf{100 rounds}}                                                                                      \\ 
\cmidrule(lr){3-4} \cmidrule(lr){5-6}
\textbf{Datasets}      &  \textbf{Epochs}                     & \textbf{FedAvg} & \textbf{FedCurv}  & \textbf{FedAvg} & \textbf{FedCurv}  \\ 
\midrule
\multirow{3}{*}{MNIST}        & 1                    & 96.33$\%$               & \textbf{96.72$\%$}     & 96.84$\%$      & \textbf{97.24$\%$}    \\
                           & 10                   & 99.07$\%$               & \textbf{99.15$\%$}     & 99.37$\%$   &   \textbf{99.40$\%$}     \\
                           & 30                  & \textbf{99.48$\%$}              & 99.38$\%$     & \textbf{99.49$\%$}      & 99.42$\%$            \\ 
\midrule
\multirow{3}{*}{CIFAR10}        & 1                    & \textbf{56.24$\%$}               & 55.13$\%$     & \textbf{56.10$\%$}      & 54.73$\%$    \\
                           & 10                   & \textbf{70.49$\%$}               & 70.35$\%$     & \textbf{74.67$\%$}      & 73.46$\%$     \\
                           & 30                  & \textbf{74.67$\%$}              & 73.42$\%$     & \textbf{76.91$\%$}      & 76.24$\%$             \\ 
\midrule
\multirow{3}{*}{MedMNIST}        & 1                    & \textbf{80.42$\%$}               & 79.44$\%$     & 83.24$\%$      & \textbf{84.28$\%$}    \\
                           & 10                   & \textbf{87.91$\%$}               & 87.23$\%$     & \textbf{90.24$\%$}      & 88.96$\%$     \\
                           & 30                  & 89.08$\%$              & \textbf{89.19$\%$}     & 89.88$\%$      & \textbf{90.31$\%$}             \\ 
\midrule
\# best performance  & & 6 & 3 & 5 & 4 \\
\bottomrule
\end{tabular}
\end{table}

\begin{table}[]
\caption{\label{tab:prior-shift}Comparison between FedAvg and FedCurv in the prior shift setting.\\}
\label{fig:priorshift}
\begin{tabular}{crccccc} 
\toprule
\multicolumn{1}{l}{}       &           &           & \multicolumn{2}{c}{\textbf{10 rounds}}                                                               & \multicolumn{2}{c}{\textbf{100 rounds}}                                                                                      \\ 
\cmidrule(lr){4-5} \cmidrule(lr){6-7}
\textbf{Category} & \textbf{Datasets}      &  \textbf{Epochs}                     & \textbf{FedAvg} & \textbf{FedCurv}  & \textbf{FedAvg} & \textbf{FedCurv}  \\ 
\midrule
\multirow{6}{*}{Labels Quantity Skew} & 
\multirow{3}{*}{CIFAR10}        & 1                    & \textbf{41.36$\%$}               & 39.93$\%$     & \textbf{47.60$\%$}      & 37.06$\%$    \\
                           & & 10                   & \textbf{45.76$\%$}               & 39.77$\%$    & \textbf{57.65$\%$}      & 46.08$\%$     \\
                           & & 30                  & \textbf{41.48$\%$}              & 26.91$\%$     & \textbf{62.52$\%$}      & 39.65$\%$             \\ 
\cmidrule{2-7} & 
\multirow{3}{*}{MedMNIST}        & 1                    & \textbf{53.64$\%$}               & 48.75$\%$     & \textbf{61.82$\%$}      & 54.01$\%$    \\
                           & & 10                   & \textbf{46.03$\%$}               & 38.98$\%$     & 60.03$\%$      & \textbf{65.68$\%$}     \\
                           & & 30                  & 37.50$\%$              & \textbf{53.90$\%$}     & \textbf{58.12$\%$}      & 56.65$\%$            \\ 
\bottomrule

\multirow{6}{*}{Dirichlet Labels Skew} & 
\multirow{3}{*}{CIFAR10}        & 1                    & \textbf{48.68$\%$}               & 47.70$\%$     & \textbf{48.50$\%$}      & 48.37$\%$    \\
                           & & 10                   & 62.39$\%$               & \textbf{62.77$\%$}    & 70.44$\%$      & \textbf{71.50$\%$}     \\
                           & & 30                  & \textbf{67.10$\%$}              & 67.01$\%$     & \textbf{75.54$\%$}      & 74.83$\%$             \\ 
\cmidrule{2-7} & 
\multirow{3}{*}{MedMNIST}        & 1                    & \textbf{76.27$\%$}               & 76.25$\%$     & \textbf{81.62$\%$}      & 81.27$\%$    \\
                           & & 10                   & \textbf{85.54$\%$}               & 84.92$\%$     & 88.74$\%$      & \textbf{89.40$\%$}     \\
                           & & 30                  & 86.92$\%$              & \textbf{86.93$\%$}     & 90.49$\%$      & \textbf{90.64$\%$}            \\ 
\bottomrule

\multirow{6}{*}{Pathological Labels Skew} & 
\multirow{3}{*}{CIFAR10}        & 1                    & \textbf{40.52$\%$}               & 35.20$\%$     & \textbf{47.42$\%$}      & 45.60$\%$    \\
                           & & 10                   & 53.52$\%$               & \textbf{60.57$\%$}    & 64.05$\%$      & \textbf{64.33$\%$}     \\
                           & & 30                  & 48.09$\%$              & \textbf{59.62$\%$}     & 61.58$\%$      & \textbf{67.09$\%$}             \\ 
\cmidrule{2-7} & 
\multirow{3}{*}{MedMNIST}        & 1                    & \textbf{64.42$\%$}               & 60.93$\%$     & \textbf{72.27$\%$}      & 70.52$\%$    \\
                           & & 10                   & \textbf{65.83$\%$}               & 57.13$\%$     & \textbf{78.81$\%$}      & 71.23$\%$     \\
                           & & 30                  & \textbf{80.36$\%$}              & 74.65$\%$     & 83.41$\%$      & \textbf{84.95$\%$}            \\ 
\midrule
\# best performance  & & & 13 & 5 & 11 & 7 \\
\bottomrule
\end{tabular}
\end{table}

\begin{table}
\caption{\label{tab:covariate-shift}Comparison between FedAvg and FedCurv in the covariate shift setting.\\}
\label{fig:covariate}
\begin{tabular}{crcccc} 
\toprule
\multicolumn{1}{l}{}       &                      & \multicolumn{2}{c}{\textbf{10 rounds}}                                                               & \multicolumn{2}{c}{\textbf{100 rounds}}                                                                                      \\ 
\cmidrule(lr){3-4} \cmidrule(lr){5-6}
\textbf{Datasets}      &  \textbf{Epochs}                     & \textbf{FedAvg} & \textbf{FedCurv}  & \textbf{FedAvg} & \textbf{FedCurv}  \\ 
\midrule
\multirow{3}{*}{CIFAR10}        & 1                    & \textbf{46.82$\%$}               & 43.55$\%$     & \textbf{48.90$\%$}      & 45.32$\%$    \\
                           & 10                   & 62.57$\%$               & \textbf{63.28$\%$}     & 66.42$\%$      & \textbf{68.90$\%$}     \\
                           & 30                  & \textbf{69.04$\%$}              & 64.56$\%$     & \textbf{73.14$\%$}      & 71.28$\%$             \\ 
\midrule
\multirow{3}{*}{MedMNIST}        & 1                    & 75.07$\%$               & \textbf{77.70$\%$}     & 79.34$\%$      & \textbf{82.32$\%$}    \\
                           & 10                   & 84.46$\%$               & \textbf{84.59$\%$}     & 87.16$\%$      & \textbf{88.18$\%$}     \\
                           & 30                  & \textbf{85.26$\%$}              & 84.99$\%$     & 89.41$\%$      & \textbf{89.45$\%$}             \\ 
\midrule
\# best performance  & & 3 & 3 & 2 & 4 \\
\bottomrule
\end{tabular}
\end{table}

The results can be analyzed from different points of view, showing interesting insights:

\noindent \textbf{Epochs per round perspective}: It can be noted that the accuracy increases as the number of epochs per round \textit{E} increases. This means that local optima may be close to the global optima, and so, with the same amount of rounds, training for more epochs can be beneficial. This pattern is clearly shown for each setting and algorithm, except for the FedAvg on MedMNIST in the uniform setting (Table \ref{tab:uniform-setting}) and in the case of the Labels Quantity Skew (Table \ref{tab:prior-shift}).

\noindent \textbf{Distribution perspective}: prior shift (see \cref{tab:prior-shift}) is the most challenging non-IID setting. In particular, among the different types of prior shift analyzed, the labels quantity skew is the most detrimental. Both FedAvg and FedCurv perform worse on the labels quantity skew. By design, the pathological labels skew is the most similar to the labels quantity skew because, in both cases, each client has examples that belong to only a small subset of the possible classes. Indeed, the pathological labels skew is the second hardest scenario after the labels quantity skew. The Dirichlet labels skew is the less challenging prior shift case. FedAvg and FedCurv perform well on the quantity skew setting (see \cref{tab:quantity-skew}). This is reasonable because both algorithms adopt a weighted averaging of the parameters, and the distribution of the examples (except for the quantity of the examples) is uniform among parties, which is the easiest setting and the one that is more similar to the non-federated scenario. The covariate shift (see ~\cref{tab:covariate-shift}) presents only a low loss of accuracy.
    
\noindent \textbf{Algorithm perspective}: despite FedCurv was born for tackling non-IID data in FL, it obtains better results than FedAvg in the uniform (Table \ref{tab:uniform-setting}) and in the covariate shift settings (Table \ref{tab:covariate-shift}). FedAvg performs well on quantity skew and prior shift scenarios, confirming that it can work on non-IID data \cite{li2019convergence}. It is interesting to note that most of the time, FedCurv wins after 100 rounds, showing that it may require more training steps to convergence.


\noindent \textbf{Communication perspective}: it seems that, with the same amount of epochs, less communication achieves better results. For example, in each setting (apart from the labels quantity skew case), after ten rounds of ten epochs, i.e. 100 epochs, both FedAvg and FedCurv have better accuracy than 100 rounds of one epoch. This means that, perhaps counter-intuitively, training locally before performing aggregation can boost the model's accuracy. This seems to indicate that pursuing local optimizations can lead to better approximations of the local optima. Why this is the case is an interesting avenue for future investigation.

\section{Conclusions}
\label{sec:conclusions}
In this paper, we experimented with two federated Learning algorithms in five different non-IID settings. In our experiments, neither of the two algorithms outperforms the other in all the partitioning strategies. However, somewhat unexpectedly, FedAvg produced better models in a majority of non-IID settings despite competing with an algorithm that was explicitly developed to improve in this scenario. Interestingly, both algorithms seem to perform better when the number of epochs per round is increased (which also has the benefit of reducing the communication cost). This is, to the best of our knowledge, a new observation, and we aim to investigate it in the future.
%
%
Among the datasets we tested, the ones implementing the quantity and pathological labels skews are those posing the hardest challenges to the algorithms. Also, as expected, the quantity skew appears to be the less challenging type of skew.
%
%
In future work, we aim to collect further datasets and expand the number of FL algorithms we test to provide a comprehensive picture of the state of the art of FL in non-IID settings. 


\section*{Acknowledgements}
This was has been partially supported by the \emph{European PILOT} project, which has received funding from the EuroHPC JU under grant agreement No.101034126. The JU receives support from the European Union’s Horizon 2020 research and innovation programme and Spain, Italy, Switzerland, Germany, France, Greece, Sweden, Croatia and Turkey.

\bibliography{sample-ceur}

\end{document}